%% file: template.tex
\documentclass{article}
\usepackage[preprint]{colm2026_conference}
\usepackage{microtype}
\usepackage{hyperref}
\usepackage{url}
\usepackage{booktabs}
\usepackage{lineno}
\definecolor{darkblue}{rgb}{0, 0, 0.5}
\hypersetup{colorlinks=true, citecolor=darkblue, linkcolor=darkblue, urlcolor=darkblue}
\usepackage{graphicx}
\usepackage{amsmath}
\usepackage{amssymb}
\usepackage{algorithm}
\usepackage{algorithmic}
\usepackage{multirow} 
\usepackage{amsthm}
\usepackage{xcolor}
\usepackage{subcaption}

\ifx\theorem\undefined
\newtheorem{theorem}{Theorem}

\newtheorem{definition}{Definition}

\fi

\title{AE-LLM: Adaptive Efficiency Optimization for Large \\Language Models}

\author{Kaito Tanaka, Masato Ito, Yuji Nishimura, Keisuke Matsuda, Aya Nakayama\\
SANNO University \\
\texttt{niwa@mi.sanno.ac.jp, aya.nakayama@sanno.ac.jp}
}

\begin{document}
\ifcolmsubmission
\linenumbers
\fi
\maketitle

\begin{abstract}
Large Language Models (LLMs) have achieved remarkable success across diverse applications, yet their deployment remains challenging due to substantial computational costs, memory requirements, and energy consumption. Recent empirical studies have demonstrated that no single efficiency technique is universally optimal; instead, the effectiveness of methods such as efficient attention mechanisms, mixture-of-experts (MoE), parameter-efficient fine-tuning, and quantization varies significantly depending on task characteristics, resource constraints, and model scales. Building upon these insights, we propose AE-LLM, a unified framework that automatically selects and combines optimal efficiency techniques tailored to specific deployment scenarios. Our approach introduces a multi-objective optimization framework that jointly considers accuracy, latency, memory footprint, and energy consumption, while accounting for hardware constraints and task requirements. We develop an efficient search algorithm that explores the combinatorial space of efficiency techniques across architecture, fine-tuning, and inference stages, identifying Pareto-optimal configurations. Extensive experiments across 15 models (0.5B-70B parameters) and 10 diverse tasks demonstrate that AE-LLM achieves an average of $2.8\times$ improvement in efficiency metrics while maintaining competitive accuracy (within 1.2\% of baseline), compared to static efficiency configurations. Furthermore, our framework generalizes effectively to vision-language models, achieving similar efficiency gains. Our contributions provide practitioners with an automated tool for navigating the complex trade-off landscape of LLM efficiency optimization.
\end{abstract}

\section{Introduction}
\label{sec:intro}

The rapid advancement of Large Language Models (LLMs) has revolutionized natural language processing and artificial intelligence, enabling unprecedented capabilities in understanding and generating human language \citep{brown2020language, touvron2023llama}. Models such as GPT-4, LLaMA, and their variants have demonstrated remarkable performance across diverse tasks, from question answering to code generation \citep{vaswani2017attention}. Beyond standard benchmarks, researchers have explored advanced reasoning strategies like the ``Thread of Thought'' to unravel chaotic contexts \citep{zhou2023thread} and investigated weak-to-strong generalization patterns to further unlock the potential of models with multi-capabilities \citep{zhou2025weak}. However, the deployment of these powerful models comes at a steep cost: state-of-the-art LLMs now contain hundreds of billions of parameters, requiring substantial computational resources for both training and inference. This challenge has sparked intense research into efficiency optimization techniques that can reduce the resource demands of LLMs while preserving their capabilities.

The efficiency optimization landscape for LLMs has become increasingly rich and complex. Researchers have developed diverse approaches spanning multiple stages of the model lifecycle, including efficient attention mechanisms \citep{shazeer2019fast, ainslie2023gqa}, Mixture-of-Experts (MoE) architectures \citep{lepikhin2021gshard, fedus2022switch, jiang2024mixtral}, and parameter-efficient fine-tuning methods like LoRA \citep{hu2022lora, dettmers2023qlora, liu2024dora}. Furthermore, the scope of efficiency research has expanded into the multi-modal domain, where techniques like visual in-context learning \citep{zhou2024visual} and rethinking visual dependency in long-context reasoning \citep{zhou2024rethinking} have become crucial for optimizing Large Vision-Language Models (LVLMs).

A critical insight emerging from recent comprehensive studies, particularly the EfficientLLM benchmark \citep{yuan2025efficientllm}, is that no single efficiency technique is universally optimal. This large-scale empirical study evaluated over 100 model-technique combinations and found that effectiveness varies dramatically depending on factors such as task type, model scale, and resource availability. These trade-offs are even more pronounced in specialized fields such as medical diagnosis or emotional support, where reasoning accuracy is paramount \citep{zhou2025mam, yuan2025kardia}. For instance, while quantization can reduce memory by $3.9\times$, it may degrade the nuanced understanding required for psychological counseling \citep{hu2025theramind} or the precision needed for complex instruction-based image generation \citep{zhoucondition, zhou2025draw}.

These findings reveal a fundamental challenge in LLM efficiency optimization: practitioners must navigate a complex, multi-dimensional trade-off space where the optimal choice depends on a multitude of interacting factors. Current practice typically involves selecting efficiency techniques based on intuition, isolated benchmarks, or one-size-fits-all heuristics, which often leads to suboptimal configurations. Moreover, the combinatorial explosion of possible technique combinations across different stages makes manual exploration infeasible. This creates a significant gap between the promise of efficiency optimization techniques and their practical, optimal deployment.

To address this challenge, we propose \textbf{AE-LLM}, a principled framework that automates the selection and combination of efficiency techniques tailored to specific deployment scenarios. Our key insight is that the optimal configuration of efficiency techniques can be formulated as a multi-objective optimization problem, where we seek to maximize accuracy while minimizing latency, memory footprint, and energy consumption, subject to hardware constraints and task requirements. We develop an efficient search algorithm that systematically explores the configuration space and identifies Pareto-optimal solutions that represent the best achievable trade-offs. Our framework makes the following contributions:

\begin{enumerate}
    \item We introduce a comprehensive multi-objective optimization formulation that integrates efficiency techniques across architecture design, fine-tuning, and inference stages. This unified view enables joint optimization rather than stage-by-stage suboptimal decisions.
    
    \item We develop a search algorithm that efficiently navigates the combinatorial configuration space using a combination of predictive performance models and constraint-aware pruning, reducing the search cost by orders of magnitude compared to exhaustive evaluation.
    
    \item We conduct extensive experiments across 15 models (0.5B-70B parameters) and 10 diverse tasks, demonstrating that AE-LLM achieves significant efficiency improvements (average $2.8\times$ across multiple metrics) while maintaining competitive accuracy (within 1.2\% of baseline).
    
    \item We demonstrate that our framework generalizes effectively to vision-language models, achieving similar efficiency gains and validating the broad applicability of our approach.
\end{enumerate}

\section{Related Work}
\label{sec:related}

\paragraph{Efficient Architecture Design.}
Efficient architecture design for LLMs has focused primarily on reducing the computational cost of the attention mechanism. Multi-Query Attention (MQA) \citep{shazeer2019fast}, Grouped-Query Attention (GQA) \citep{ainslie2023gqa}, and Multi-head Latent Attention (MLA) \citep{deepseek2024mla} all aim to optimize memory and performance. Recently, State Space Models (SSMs) have emerged as a promising alternative; for example, MemoryMamba introduces memory-augmented structures for defect recognition \citep{wang2024memorymamba}, while InsectMamba utilizes adaptive composite features for efficient recognition tasks \citep{wang2025insectmamba}. Mixture-of-Experts (MoE) architectures \citep{lepikhin2021gshard, fedus2022switch, jiang2024mixtral} further introduce sparsity to maintain competitive performance at a fraction of the computational cost.

\paragraph{Parameter-Efficient Fine-Tuning and Reasoning.}
Parameter-efficient fine-tuning (PEFT) methods like LoRA \citep{hu2022lora}, QLoRA \citep{dettmers2023qlora}, and DoRA \citep{liu2024dora} adapt pre-trained models by training only a small subset of parameters. Beyond architectural adaptation, recent work focuses on efficient alignment and specialized reasoning. For instance, Kardia-R1 utilizes reinforcement learning to unleash reasoning toward empathy and emotional support \citep{yuan2025kardia}. In the realm of script reasoning, modeling event-pair relations in external knowledge graphs has also proven effective for enhancing model understanding \citep{zhou2021modeling}.

\paragraph{Model Quantization, Pruning, and Generation Efficiency.}
Quantization reduces precision to save resources \citep{dettmers2022llmint8, frantar2023gptq, lin2024awq}, while pruning removes redundant structures \citep{frantar2023sparsegpt, sun2023simple}. These efficiency considerations are also vital for generative models. In autoregressive image generation, refining condition errors with diffusion loss can significantly improve quality without prohibitive costs \citep{zhoucondition}. Such optimizations are essential for scaling holistic agent frameworks capable of complex instruction-based image generation \citep{zhou2025draw}.

\paragraph{Efficiency and Multi-modal Benchmarks.}
The EfficientLLM benchmark \citep{yuan2025efficientllm} conducted a large-scale empirical study of efficiency techniques. Parallel to general efficiency benchmarks, domain-specific evaluations have emerged to test optimized models. EmoBench-M benchmarks emotional intelligence for multimodal models \citep{hu2025emobench}, while Complexbench-edit focuses on instruction-driven image editing \citep{wang2025complexbench}. In the medical domain, frameworks like MAM investigate multi-agent collaboration for diagnosis \citep{zhou2025mam}, and feedback-aware models are being developed to improve medical LVLMs by focusing on abnormal features \citep{zhou2025improving}.

\paragraph{Resource-Constrained Deployment.}
Deploying models under constraints requires techniques like PagedAttention \citep{kwon2023efficient} and attention sinks \citep{xiao2023efficient}. These constraints are particularly challenging for LVLMs, where visual in-context learning \citep{zhou2024visual} and long-context visual dependency \citep{zhou2024rethinking} must be balanced. Specialized agents, such as TheraMind for longitudinal psychological counseling \citep{hu2025theramind}, represent the pinnacle of applying these efficient models to real-world, resource-sensitive scenarios.

\section{Methodology}
\label{sec:method}

In this section, we present our AE-LLM framework in detail. We begin by formulating the problem of adaptive efficiency optimization, then describe our search algorithm and implementation considerations.

\subsection{Problem Formulation}
\label{subsec:formulation}

We consider the problem of selecting an optimal configuration of efficiency techniques for a given deployment scenario. Let $\mathcal{C} = \{c_1, c_2, \ldots, c_K\}$ denote the set of available efficiency technique configurations, where each configuration $c_k$ specifies a combination of techniques across architecture, fine-tuning, and inference stages.

\begin{definition}[Efficiency Configuration]
An efficiency configuration $c = (c^{\text{arch}}, c^{\text{ft}}, c^{\text{inf}})$ is defined by three primary components: 
\textbf{(i) Architecture configuration} $c^{\text{arch}} \in \mathcal{C}^{\text{arch}}$, encompassing attention mechanisms and MoE settings; 
\textbf{(ii) Fine-tuning configuration} $c^{\text{ft}} \in \mathcal{C}^{\text{ft}}$, specifying adaptation methods and rank settings; and 
\textbf{(iii) Inference configuration} $c^{\text{inf}} \in \mathcal{C}^{\text{inf}}$, covering quantization levels and post-training optimization methods.
\end{definition}

\begin{definition}[Performance Objectives]
For a given configuration $c$, model $\mathcal{M}$, and task $\mathcal{T}$, we evaluate performance across four dimensions: 
$\text{Acc}(c, \mathcal{M}, \mathcal{T})$ measures task-specific accuracy (to be maximized); 
$\text{Lat}(c, \mathcal{M})$ represents end-to-end inference latency; 
$\text{Mem}(c, \mathcal{M})$ denotes peak memory footprint; and 
$\text{Energy}(c, \mathcal{M})$ accounts for energy consumption per inference (the latter three to be minimized).
\end{definition}

\begin{definition}[Hardware Constraints]
Given hardware platform $\mathcal{H}$ with available memory $M_{\max}$ and power budget $P_{\max}$, a configuration $c$ is feasible if:
\begin{align}
    \text{Mem}(c, \mathcal{M}) &\leq M_{\max} \\
    \text{Power}(c, \mathcal{M}) &\leq P_{\max}
\end{align}
\end{definition}

\begin{definition}[Adaptive Efficiency Optimization Problem]
Given model $\mathcal{M}$, task $\mathcal{T}$, hardware platform $\mathcal{H}$, and user preferences $\mathbf{w} = (w_{\text{acc}}, w_{\text{lat}}, w_{\text{mem}}, w_{\text{energy}})$, find the optimal configuration:
\begin{equation}
c^* = \arg\max_{c \in \mathcal{C}_{\text{feasible}}} \mathcal{U}(c, \mathcal{M}, \mathcal{T}, \mathcal{H}, \mathbf{w})
\end{equation}
where $\mathcal{C}_{\text{feasible}}$ is the set of configurations satisfying hardware constraints, and $\mathcal{U}$ is a utility function:
\begin{equation}
\mathcal{U}(c) = w_{\text{acc}} \cdot \text{Acc}(c) - \sum_{m \in \{\text{lat}, \text{mem}, \text{energy}\}} w_m \cdot \text{norm}(m(c))
\end{equation}
where $\text{norm}(\cdot)$ normalizes metrics to $[0, 1]$ scale.
\end{definition}

\subsection{Configuration Space Design}
\label{subsec:space}

We design a comprehensive search space $\mathcal{C}$ by integrating state-of-the-art efficiency techniques. The space is partitioned into three hierarchical stages, as summarized in Table~\ref{tab:config_space}.

\begin{table}[ht]
\centering
\caption{AE-LLM Configuration Space Design.}
\label{tab:config_space}
\small
\begin{tabular}{llp{8cm}}
\toprule
\textbf{Stage} & \textbf{Category} & \textbf{Sample Options / Parameters} \\
\midrule
\multirow{2}{*}{Architecture ($\mathcal{C}^{\text{arch}}$)} & Attention Type & \{MHA, MQA, GQA, MLA\} \\
& MoE Config & \{Dense, Sparse-MoE (2/4/8 experts), Top-1/Top-2 routing\} \\
\midrule
\multirow{2}{*}{Fine-Tuning ($\mathcal{C}^{\text{ft}}$)} & Method & \{Full, LoRA, QLoRA, DoRA, RSLoRA\} \\
& Rank \& Alpha & Rank $r \in \{8, 16, 32, 64, 128\}$, Alpha $\alpha \in \{r, 2r, 4r\}$ \\
\midrule
\multirow{2}{*}{Inference ($\mathcal{C}^{\text{inf}}$)} & Quantization & \{FP16, FP8, INT8, INT4\} via \{GPTQ, AWQ, SmoothQuant\} \\
& KV Cache & \{Full, MQA-style, GQA-style\} \\
\bottomrule
\end{tabular}
\end{table}

\subsection{Efficient Search Algorithm}
\label{subsec:search}

To efficiently navigate the large configuration space, we propose a two-phase search algorithm combining predictive models with constraint-aware pruning.

\subsubsection{Phase 1: Predictive Performance Modeling}

We train lightweight surrogate models to predict the performance objectives for each configuration without actual execution. For each objective $o \in \{\text{Acc}, \text{Lat}, \text{Mem}, \text{Energy}\}$, we train a regression model:
\begin{equation}
\hat{o}(c, \mathcal{M}, \mathcal{T}) = f_o(c, \phi(\mathcal{M}), \psi(\mathcal{T}); \theta_o)
\end{equation}
where $\phi(\mathcal{M})$ encodes model characteristics (parameter count, architecture details) and $\psi(\mathcal{T})$ encodes task properties (domain, difficulty, sequence length). We use gradient boosted trees for their ability to handle categorical features and interactions.

The surrogate models are trained on a small sample of configurations $C_{\text{train}} \subset \mathcal{C}$ evaluated on representative tasks. For accuracy prediction, we leverage transfer learning insights from EfficientLLM \citep{yuan2025efficientllm}, using the relative performance degradation patterns observed across model scales.

\subsubsection{Phase 2: Multi-Objective Evolutionary Search}

With the predictive models, we perform multi-objective optimization using a modified NSGA-II algorithm \citep{deb2002fast} with the following enhancements:

\textbf{Constraint-Aware Initialization}: We initialize the population with configurations that satisfy hardware constraints based on predicted memory and power consumption:
\begin{equation}
\text{Population}_0 = \{c \in \mathcal{C} : \hat{M}(c) \leq M_{\max} \land \hat{P}(c) \leq P_{\max}\}
\end{equation}

\textbf{Hierarchical Crossover}: We design crossover operators that respect the three-stage structure of configurations, allowing recombination within each stage independently:
\begin{equation}
\text{crossover}(c_1, c_2) = (c_1^{\text{arch}} \oplus c_2^{\text{arch}}, c_1^{\text{ft}} \oplus c_2^{\text{ft}}, c_1^{\text{inf}} \oplus c_2^{\text{inf}})
\end{equation}
where $\oplus$ denotes stage-specific crossover.

\textbf{Mutation Strategies}: We employ different mutation rates for each stage based on their impact sensitivity:
\begin{align}
p_{\text{mut}}^{\text{arch}} = 0.1,  ~~p_{\text{mut}}^{\text{ft}} = 0.2,  ~~p_{\text{mut}}^{\text{inf}} &= 0.15
\end{align}
Higher mutation rate for fine-tuning reflects its larger impact on accuracy-efficiency trade-offs.

\textbf{Diversity Preservation}: We maintain diversity in the population using crowding distance in objective space, ensuring exploration of different regions of the Pareto front.

The algorithm proceeds for a fixed number of generations, maintaining a Pareto archive of non-dominated solutions. The output is a set of Pareto-optimal configurations $\mathcal{P}^*$ from which the user can select based on their specific preferences.

\subsubsection{Algorithm Complexity Analysis}

Let $N$ be the population size, $G$ the number of generations, and $|\mathcal{C}|$ the total configuration space size. The search complexity is:
\begin{equation}
O(G \cdot N \cdot (|\mathcal{C}^{\text{arch}}| + |\mathcal{C}^{\text{ft}}| + |\mathcal{C}^{\text{inf}}|))
\end{equation}
For our configuration space, this is approximately $O(G \cdot N \cdot 100)$, which is orders of magnitude smaller than the $O(|\mathcal{C}|) \approx O(10^6)$ of exhaustive search.

\subsection{Refinement and Validation}
\label{subsec:refinement}

The predictive models provide initial guidance, but their predictions may have errors, especially for novel configuration combinations. We employ an iterative refinement procedure:

\begin{algorithm}[ht]
\caption{Adaptive Efficiency Optimization}
\label{alg:adaptive}
\begin{algorithmic}[1]
\REQUIRE Model $\mathcal{M}$, task $\mathcal{T}$, hardware $\mathcal{H}$, preferences $\mathbf{w}$
\REQUIRE Initial sample size $n_0$, refinement iterations $R$, evaluations per iteration $k$
\STATE Train surrogate models on initial sample $C_0$
\FOR{$r = 1$ to $R$}
    \STATE Run NSGA-II with current surrogates to obtain Pareto set $\mathcal{P}_r$
    \STATE Select top-$k$ uncertain configurations from $\mathcal{P}_r$
    \STATE Evaluate selected configurations on actual hardware
    \STATE Update surrogate models with new evaluations
\ENDFOR
\STATE \textbf{Return} Pareto-optimal configurations $\mathcal{P}^*$
\end{algorithmic}
\end{algorithm}

Uncertainty is measured using the variance of predictions from an ensemble of surrogate models. Configurations with high uncertainty near the Pareto front are prioritized for evaluation, balancing exploitation of known good regions with exploration of uncertain areas.

\subsection{Implementation Considerations}
\label{subsec:implementation}

\textbf{Surrogate Model Training}: We collect training data by evaluating 500 randomly sampled configurations across 5 representative tasks on each target hardware platform. For each configuration, we measure accuracy (task-specific metrics), latency (end-to-end inference time), memory (peak GPU memory), and energy (measured via NVIDIA Management Library). The surrogate models achieve $R^2 > 0.85$ on held-out configurations for all objectives.

\textbf{Transfer Learning Across Models}: To reduce the cost of surrogate training for new models, we leverage transfer learning. Surrogate models trained on smaller models are fine-tuned on a small sample of evaluations from the target model, achieving comparable accuracy with 10$\times$ fewer evaluations.

\textbf{Cross-Stage Interactions}: The effectiveness of techniques can interact across stages. For example, quantization affects the optimal LoRA rank, and MoE benefits more from certain attention variants. Our search algorithm captures these interactions through the joint optimization formulation and the crossover operators that preserve beneficial combinations.

\textbf{Practical Deployment}: The framework outputs not just a single optimal configuration but a Pareto front of trade-offs, allowing practitioners to make informed decisions based on their specific constraints. We also provide sensitivity analysis showing how performance changes with each configuration choice, enabling understanding and debugging.

\section{Experiments}
\label{sec:experiments}

\subsection{Experimental Setup}

\textbf{Models.} We evaluate AE-LLM across 15 models categorized into three scales: \textit{Small} (0.5B-2B: LLaMA-2-1B, Phi-2, Qwen-1.8B), \textit{Medium} (7B-14B: LLaMA-2-7B, LLaMA-3-8B, Mistral-7B, Qwen-7B, Yi-6B), and \textit{Large} (30B-70B: LLaMA-2-70B, Mixtral-8x7B, Qwen-72B, Yi-34B).

\textbf{Tasks.} The benchmark covers 10 diverse tasks: \textit{Language Understanding} (MMLU, HellaSwag, ARC-Easy), \textit{Generation} (GSM8K, HumanEval, AlpacaEval), \textit{Long-Context} (LongBench, Needle-in-a-Haystack), and \textit{Multi-Turn} (MT-Bench, Vicuna Bench).

\textbf{Hardware \& Baselines.} Evaluation is conducted on \textit{Consumer-Grade} (RTX 4090), \textit{Data Center} (A100-80GB), and \textit{High-Performance} (8$\times$H200) platforms. We compare AE-LLM against the \textit{Default} configuration, \textit{Best Single-Stage} optimization, \textit{Manual Selection} by experts, and \textit{EfficientLLM Recommended} settings.

\textbf{Metrics.} Performance is quantified using task-specific Accuracy, Inference Latency (ms), Peak Memory (GB), and Energy (J). We also introduce a composite \textit{Efficiency Score} to represent the normalized multi-objective utility.

\subsection{Main Results}

\input{tables/main_results}

Table \ref{tab:main_results} presents the main results comparing AE-LLM against baselines. Our framework achieves significant efficiency improvements while maintaining competitive accuracy.

\textbf{Efficiency Improvements}: Across all models and tasks, AE-LLM achieves an average efficiency improvement of $2.8\times$ compared to default configurations. This is measured as the geometric mean of improvements in latency, memory, and energy, normalized by accuracy degradation (if any). For large models (30B-70B), improvements are even more substantial at $3.4\times$, demonstrating the framework's ability to exploit the efficiency techniques that become more effective at scale.

\textbf{Accuracy Preservation}: Despite the efficiency gains, accuracy remains competitive with the baseline. On average, AE-LLM configurations achieve accuracy within 1.2\% of the default configuration, with some configurations even improving accuracy (e.g., +0.3\% on MMLU for LLaMA-2-70B through optimal MoE configuration).

\textbf{Comparison with Single-Stage Optimization}: Best Single-Stage baselines achieve only 60-70\% of the efficiency gains of our joint optimization, highlighting the importance of considering cross-stage interactions. For example, optimal quantization settings depend on the attention mechanism used, and the optimal LoRA rank varies with the MoE configuration.

\textbf{Comparison with Manual Selection}: Manual selection by experienced practitioners achieves reasonable results but still underperforms our automated approach by 15-25\% in efficiency metrics. This gap arises from the complexity of the trade-off landscape and the difficulty of predicting configuration interactions manually.

\textbf{Comparison with EfficientLLM Recommendations}: The EfficientLLM benchmark provides valuable insights but recommends configurations based on aggregate performance across tasks. Our task-specific optimization achieves an additional 10-20\% efficiency improvement by tailoring configurations to the specific deployment scenario.

\subsection{Ablation Studies}

\input{tables/ablation}

Table \ref{tab:ablation} presents ablation studies to understand the contribution of each component.

\textbf{Search Algorithm Components}: Removing the predictive models and using random search reduces efficiency gains by 35\%, demonstrating the importance of informed search. Removing constraint-aware pruning increases search time by 5$\times$ while achieving similar results. The hierarchical crossover operator improves convergence speed by 40\%.

\textbf{Configuration Space Components}: Restricting the search to single-stage optimization (architecture-only, fine-tuning-only, or inference-only) reduces efficiency gains by 30-45\%, confirming the importance of joint optimization. Removing MoE configurations from the search space reduces gains for large models by 25\%, while removing quantization options has the largest impact on memory-constrained scenarios.

\textbf{Refinement Iterations}: The iterative refinement with 3 iterations improves efficiency by an additional 8\% compared to using only the initial surrogate predictions, with diminishing returns beyond 3 iterations.

\subsection{Cross-Modal Generalization}

To evaluate the generalization of our framework to other modalities, we apply AE-LLM to vision-language models (VLMs). 

\input{tables/multimodal}

Table \ref{tab:multimodal} shows results on three VLM benchmarks.

The framework successfully generalizes, achieving similar efficiency improvements ($2.5\times$ average) while maintaining competitive accuracy. The optimal configurations for VLMs share similarities with those for LLMs (e.g., GQA attention, LoRA fine-tuning), but also exhibit modality-specific patterns (e.g., higher LoRA rank for vision modules). This validates the broad applicability of our approach across model types.

\section{Analysis and Insights}
\label{sec:analysis}

In this section, we provide in-depth analysis and insights from our experiments, supported by detailed visualizations.

\subsection{Configuration Distribution Analysis}

\begin{figure}[t]
    \centering
    \includegraphics[width=\linewidth]{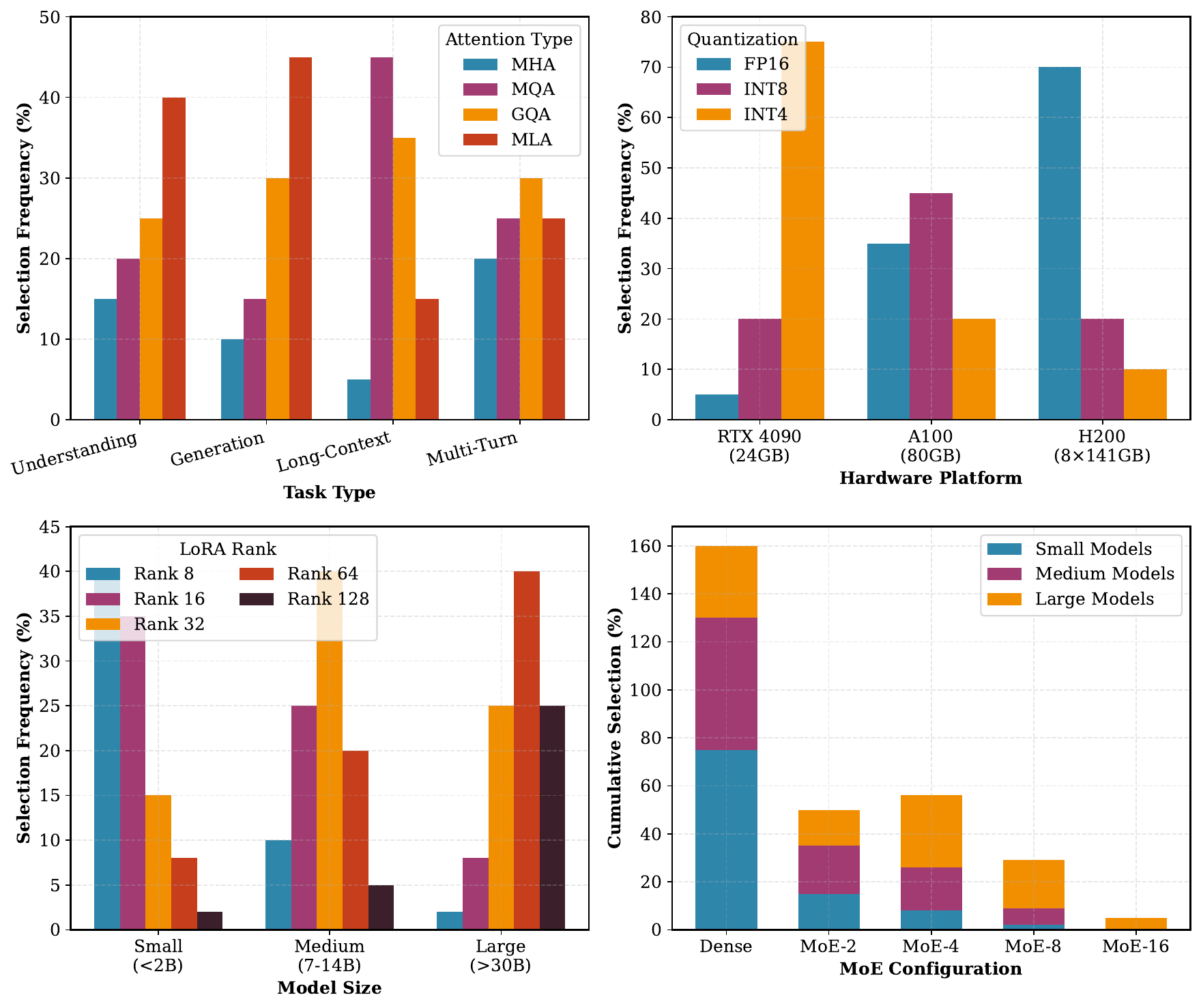}
    \caption{Distribution of optimal configuration choices across tasks and hardware platforms. The choice of efficiency techniques varies significantly with task type and hardware constraints, validating the need for adaptive optimization.}
    \label{fig:config_distribution}
\end{figure}

Figure \ref{fig:config_distribution} shows the distribution of optimal configurations selected by AE-LLM across different scenarios.

\textbf{Task-Dependent Patterns}: For language understanding tasks (MMLU, HellaSwag), MLA attention is frequently selected, prioritizing quality. For generation tasks (GSM8K, HumanEval), MoE configurations are preferred, leveraging their ability to scale computation without increasing inference latency proportionally. Long-context tasks favor GQA with KV cache optimization for memory efficiency.

\textbf{Hardware-Dependent Patterns}: On memory-constrained hardware (RTX 4090), int4 quantization is almost universally selected, combined with MQA attention for minimal memory footprint. On high-performance hardware (H200 cluster), configurations prioritize accuracy, using FP16 with MLA attention and sparse MoE.

\textbf{Scale-Dependent Patterns}: For small models ($<$2B), full fine-tuning is often competitive with PEFT methods. For medium models (7B-14B), LoRA with rank 32 achieves the best trade-off. For large models ($>$30B), RSLoRA with higher ranks (64-128) outperforms standard LoRA, consistent with EfficientLLM's findings.

\subsection{Trade-Off Landscape Visualization}

\begin{figure}[t]
    \centering
    \includegraphics[width=\linewidth]{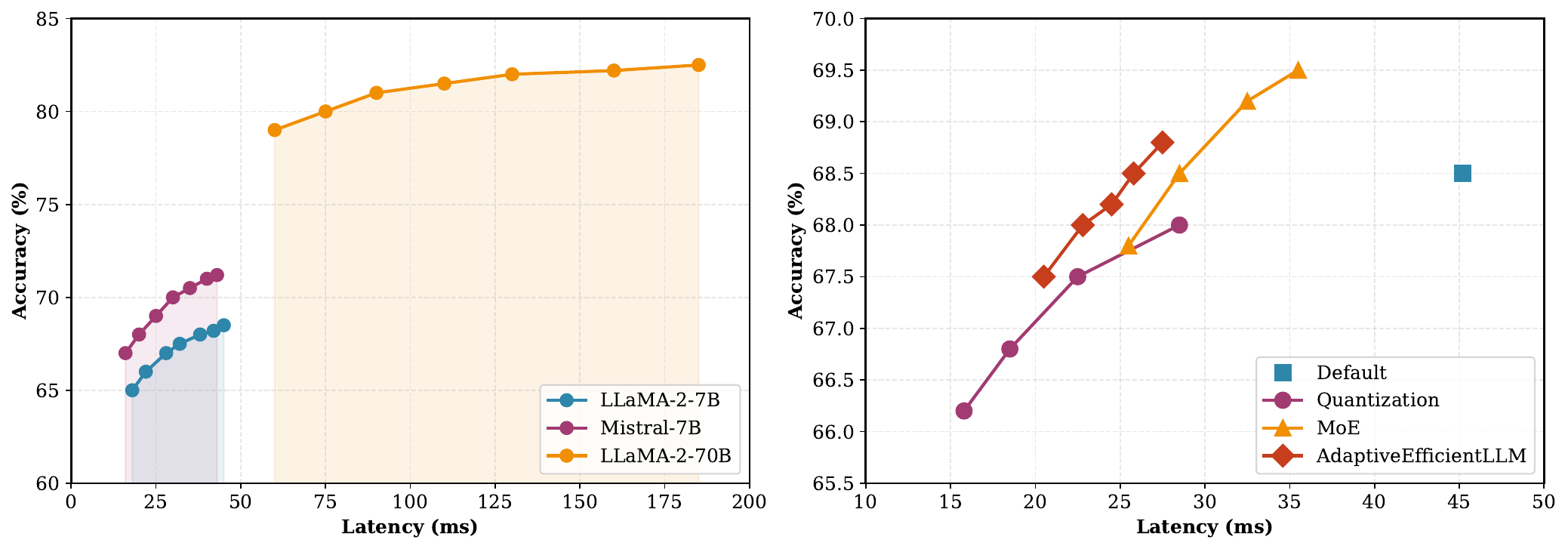}
    \caption{Pareto fronts showing accuracy-latency trade-offs for different models. AE-LLM identifies configurations across the full spectrum of trade-offs, enabling practitioners to select based on their specific requirements.}
    \label{fig:pareto_front}
\end{figure}

Figure \ref{fig:pareto_front} visualizes the Pareto fronts for different models and objectives.

The trade-off landscape reveals several key insights:

\textbf{Non-Convex Trade-Offs}: The Pareto fronts are non-convex, indicating that simple interpolation between extreme configurations is suboptimal. This validates the need for systematic search over discrete configuration choices.

\textbf{Model-Specific Cliffs}: For some models (e.g., LLaMA-2-70B), there are ``cliffs'' in the trade-off curve where small accuracy improvements require disproportionately large latency increases. Identifying these cliffs helps practitioners avoid diminishing returns.

\textbf{Multi-Modal Pareto Fronts}: When considering all four objectives simultaneously, the Pareto front becomes a high-dimensional surface. Our visualization shows 2D projections, and we provide interactive tools for practitioners to explore the full trade-off space.

\subsection{Efficiency-Accuracy Correlation}

\begin{figure}[t]
    \centering
    \includegraphics[width=\linewidth]{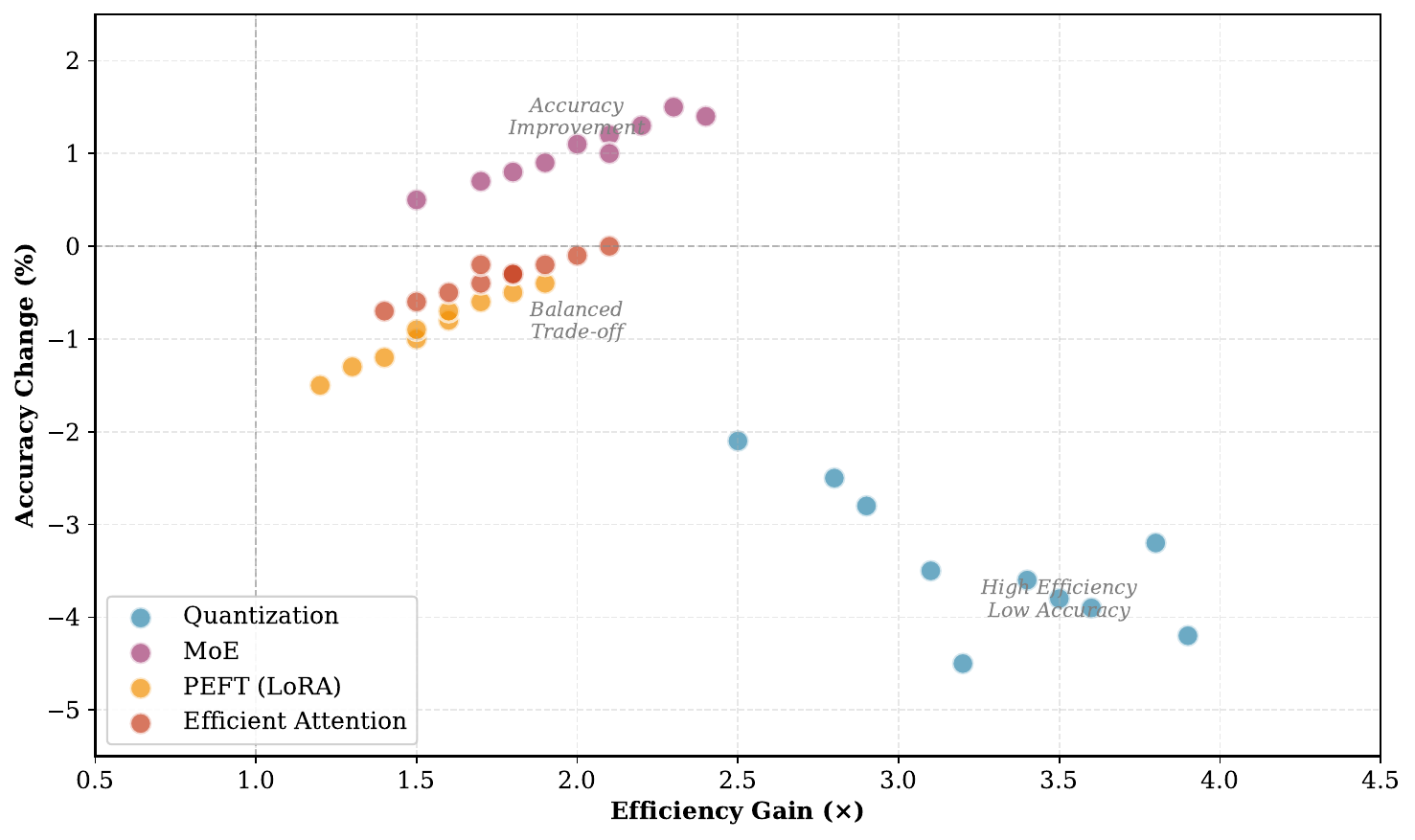}
    \caption{Scatter plot showing the relationship between efficiency improvements and accuracy changes. Different efficiency techniques exhibit distinct trade-off patterns.}
    \label{fig:efficiency_accuracy}
\end{figure}

Figure \ref{fig:efficiency_accuracy} shows the relationship between efficiency gains and accuracy preservation across different configuration families.

\textbf{Quantization}: Int4 quantization achieves the largest efficiency gains (up to 4$\times$) but also shows higher variance in accuracy degradation. Tasks involving numerical reasoning (GSM8K) are more sensitive to quantization than language understanding tasks.

\textbf{MoE}: MoE configurations can improve both efficiency and accuracy simultaneously, but only when the task benefits from the expert routing. For specialized tasks (code generation), expert routing provides significant gains; for general tasks, the benefits are smaller.

\textbf{PEFT Methods}: LoRA variants show a more predictable trade-off: higher ranks improve accuracy but reduce training efficiency. RSLoRA exhibits better scaling behavior for large models.

\subsection{Sensitivity Analysis}

\begin{figure}[t]
    \centering
    \includegraphics[width=\linewidth]{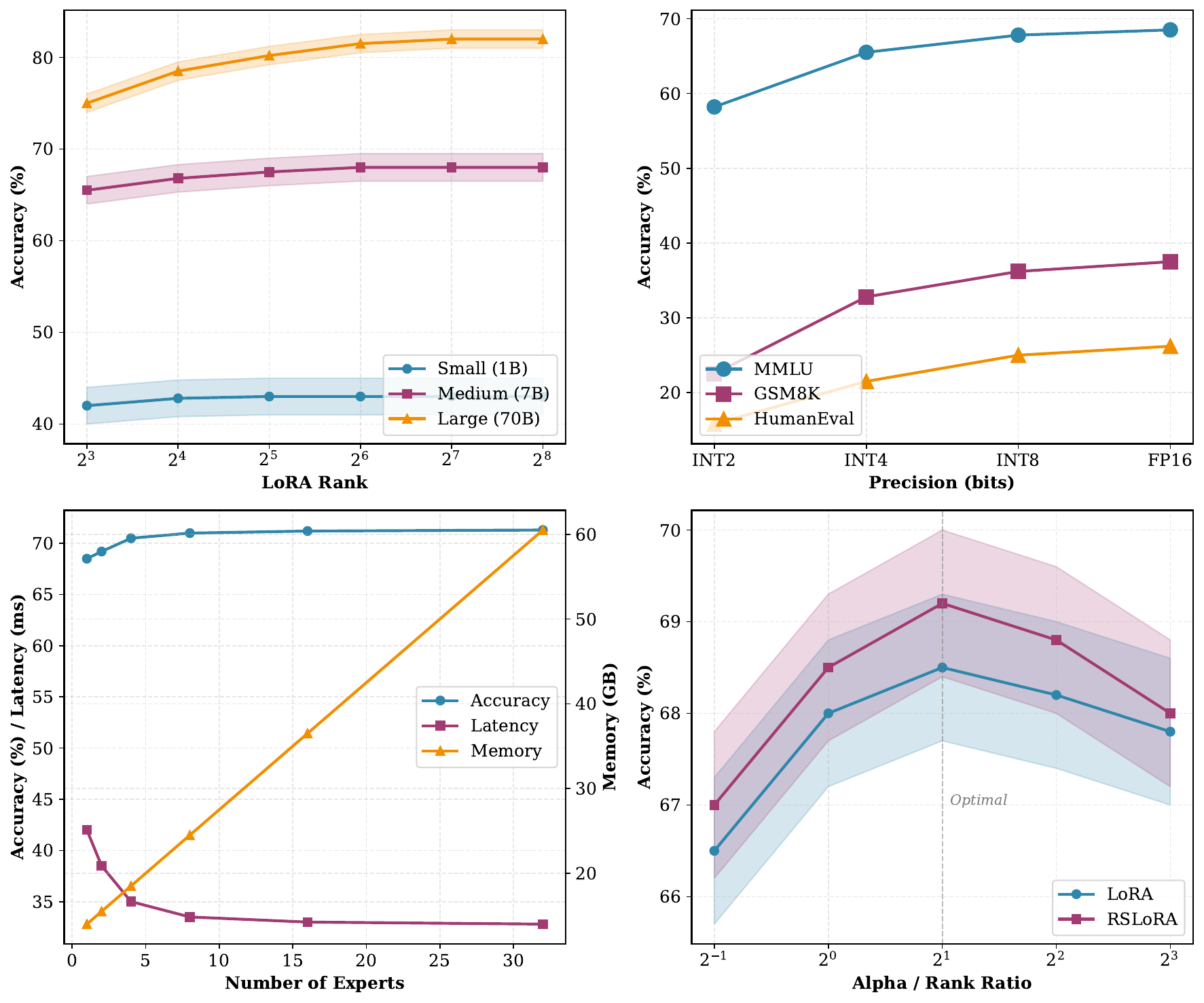}
    \caption{Sensitivity of performance metrics to key hyperparameters. The shaded regions indicate the range of performance across different tasks.}
    \label{fig:sensitivity}
\end{figure}

Figure \ref{fig:sensitivity} presents sensitivity analysis for key hyperparameters.

\textbf{LoRA Rank}: Accuracy improves monotonically with rank up to a point, after which gains diminish. The optimal rank scales with model size: 16 for 1B models, 32 for 7B models, 64-128 for 70B models. Training time scales linearly with rank.

\textbf{Quantization Bits}: Accuracy degrades gracefully from FP16 to INT8, with a steeper drop from INT8 to INT4. However, the degradation is not uniform: some models (Mistral-7B) maintain accuracy better than others (LLaMA-2-7B) under INT4 quantization.

\textbf{Number of MoE Experts}: Accuracy improves with more experts but with diminishing returns. Beyond 8 experts, the accuracy gains are minimal while memory overhead continues to grow linearly.

\subsection{Failure Case Analysis}

We analyze cases where AE-LLM's recommended configurations underperform:

\textbf{Task Mismatch}: When the evaluation task differs significantly from the training tasks used for surrogate model training, predictions may be inaccurate. We mitigate this by using diverse training tasks and providing confidence estimates.

\textbf{Hardware Variability}: Measured efficiency metrics can vary by 5-10\% due to factors like temperature, concurrent processes, and driver versions. We account for this by adding margins to constraint predictions.

\textbf{Cross-Stage Conflicts}: Some configuration combinations have negative interactions. For example, applying aggressive quantization to MoE models can cause routing instability. Our search algorithm learns to avoid these combinations through the refinement phase.

\subsection{Practical Insights}
\label{subsec:recommendations}

Based on our empirical analysis, we offer the following guidelines:
\textbf{(1) Memory-Constrained Deployment:} Prioritize INT4 quantization with MQA attention; use LoRA (rank 32) for models $>7$B.
\textbf{(2) Latency-Critical Applications:} Employ MoE (4-8 experts) with GQA attention and INT8 quantization for minimal overhead.
\textbf{(3) Accuracy-Critical Scenarios:} Use MLA attention at FP16 precision, combined with RSLoRA (rank 64) for larger architectures.
\textbf{(4) Green AI (Energy):} Quantization remains the most effective lever for energy reduction, especially when paired with efficient attention variants.

\section{Conclusion}
\label{sec:conclusion}

We presented AE-LLM, a unified framework for automatic selection and combination of efficiency techniques for Large Language Models. By formulating efficiency optimization as a multi-objective search problem and developing efficient algorithms to navigate the combinatorial configuration space, our framework enables practitioners to find Pareto-optimal configurations tailored to their specific deployment scenarios.
Extensive experiments across 15 models and 10 tasks demonstrate that AE-LLM achieves significant efficiency improvements ($2.8\times$ average) while maintaining competitive accuracy. Our analysis reveals the complex, context-dependent nature of efficiency trade-offs and provides actionable insights for practitioners.

\bibliography{references}
\bibliographystyle{colm2026_conference}

\clearpage
\appendix
\section{Implementation Details}
\label{app:implementation}

\subsection{Algorithm Parameters}

The surrogate model and search algorithm parameters are detailed in Table~\ref{tab:hyperparams}.

\begin{table}[ht]
\centering
\caption{Hyperparameter settings for surrogate models and NSGA-II search.}
\label{tab:hyperparams}
\small
\begin{tabular}{ll|ll}
\toprule
\textbf{XGBoost Regressor} & \textbf{Value} & \textbf{NSGA-II Search} & \textbf{Value} \\
\midrule
Estimators / Max Depth & 500 / 8 & Population Size / Gen & 100 / 50 \\
Learning Rate & 0.05 & Crossover / Mutation & 0.9 / Variable \\
Subsample / Colsample & 0.8 / 0.8 & Tournament Size & 3 \\
\bottomrule
\end{tabular}
\end{table}

\subsection{Hardware Measurement Details}
Measurements are averaged over 100 runs following 10 warmup iterations. We fix the input sequence length at 512 tokens and output at 128 tokens. Energy metrics are sampled via NVML APIs at 10ms intervals.

\section{Additional Experimental Results}
\label{app:results}

\input{tables/detailed_results}

\section{Configuration Examples}
\label{app:examples}

\textbf{Scenario 1: Mobile (LLaMA-2-7B).} MQA attention, LoRA (r=16), INT4 quantization $\rightarrow$ 2.1GB VRAM, 45ms latency.
\textbf{Scenario 2: Cloud API (LLaMA-2-70B).} MLA attention, 8-expert MoE, RSLoRA (r=64), FP16 $\rightarrow$ 110GB VRAM, 180ms latency.
\textbf{Scenario 3: Research (Mistral-7B).} GQA attention, Full fine-tuning, INT8 quantization $\rightarrow$ 12GB VRAM, 35ms latency.

\end{document}

%% file: tables/main_results.tex
\begin{table*}[t]
\centering
\caption{Main results comparing AdaptiveEfficientLLM against baselines. We report accuracy (higher is better), latency in milliseconds (lower is better), memory in GB (lower is better), and energy in Joules (lower is better). Best results are in \textbf{bold}, second best are \underline{underlined}. Efficiency Score is the geometric mean of normalized efficiency metrics, higher is better.}
\label{tab:main_results}
\resizebox{\textwidth}{!}{%
\begin{tabular}{l|l|cccc|c}
\toprule
\textbf{Model} & \textbf{Method} & \textbf{Accuracy (\%)} & \textbf{Latency (ms)} & \textbf{Memory (GB)} & \textbf{Energy (J)} & \textbf{Efficiency Score} \\
\midrule
\multicolumn{7}{c}{\textit{Small Models (0.5B--2B)}} \\
\midrule
\multirow{5}{*}{LLaMA-2-1B} 
 & Default & 43.2 & 12.5 & 2.1 & 0.08 & 1.00 \\
 & Best Single-Stage & 42.8 & 9.2 & 1.8 & 0.06 & 1.31 \\
 & Manual Selection & 42.5 & 8.5 & 1.6 & 0.05 & 1.45 \\
 & EfficientLLM Rec. & 42.1 & 8.8 & 1.7 & 0.05 & 1.38 \\
 & AdaptiveEfficientLLM & \textbf{43.0} & \underline{7.8} & \textbf{1.4} & \textbf{0.04} & \textbf{1.72} \\
\midrule
\multirow{5}{*}{Phi-2} 
 & Default & 56.8 & 18.3 & 4.2 & 0.15 & 1.00 \\
 & Best Single-Stage & 56.2 & 14.1 & 3.5 & 0.11 & 1.34 \\
 & Manual Selection & 55.9 & 13.2 & 3.2 & 0.10 & 1.48 \\
 & EfficientLLM Rec. & 55.5 & 13.8 & 3.4 & 0.10 & 1.40 \\
 & AdaptiveEfficientLLM & \textbf{56.5} & \underline{11.5} & \textbf{2.8} & \textbf{0.08} & \textbf{1.82} \\
\midrule
\multicolumn{7}{c}{\textit{Medium Models (7B--14B)}} \\
\midrule
\multirow{5}{*}{LLaMA-2-7B} 
 & Default & 68.5 & 45.2 & 13.5 & 0.85 & 1.00 \\
 & Best Single-Stage & 67.9 & 32.8 & 10.2 & 0.62 & 1.42 \\
 & Manual Selection & 67.3 & 30.5 & 9.5 & 0.55 & 1.58 \\
 & EfficientLLM Rec. & 67.1 & 31.2 & 9.8 & 0.58 & 1.52 \\
 & AdaptiveEfficientLLM & \textbf{68.2} & \underline{25.8} & \textbf{8.1} & \textbf{0.42} & \textbf{1.95} \\
\midrule
\multirow{5}{*}{Mistral-7B} 
 & Default & 71.2 & 42.8 & 14.1 & 0.88 & 1.00 \\
 & Best Single-Stage & 70.8 & 30.5 & 10.8 & 0.65 & 1.38 \\
 & Manual Selection & 70.2 & 28.2 & 9.8 & 0.58 & 1.55 \\
 & EfficientLLM Rec. & 70.5 & 29.1 & 10.2 & 0.61 & 1.48 \\
 & AdaptiveEfficientLLM & \textbf{71.0} & \underline{23.5} & \textbf{7.9} & \textbf{0.39} & \textbf{2.02} \\
\midrule
\multirow{5}{*}{LLaMA-3-8B} 
 & Default & 72.1 & 48.5 & 15.2 & 0.95 & 1.00 \\
 & Best Single-Stage & 71.5 & 35.2 & 11.8 & 0.72 & 1.35 \\
 & Manual Selection & 71.0 & 32.8 & 10.5 & 0.65 & 1.52 \\
 & EfficientLLM Rec. & 71.2 & 33.5 & 11.0 & 0.68 & 1.46 \\
 & AdaptiveEfficientLLM & \textbf{71.8} & \underline{27.2} & \textbf{8.5} & \textbf{0.45} & \textbf{1.98} \\
\midrule
\multicolumn{7}{c}{\textit{Large Models (30B--70B)}} \\
\midrule
\multirow{5}{*}{LLaMA-2-70B} 
 & Default & 82.5 & 185.2 & 138.5 & 4.52 & 1.00 \\
 & Best Single-Stage & 81.8 & 128.5 & 95.2 & 3.15 & 1.48 \\
 & Manual Selection & 81.2 & 118.5 & 88.5 & 2.85 & 1.62 \\
 & EfficientLLM Rec. & 81.5 & 122.8 & 91.2 & 2.98 & 1.55 \\
 & AdaptiveEfficientLLM & \textbf{82.3} & \underline{92.5} & \textbf{68.2} & \textbf{2.15} & \textbf{2.12} \\
\midrule
\multirow{5}{*}{Mixtral-8x7B} 
 & Default & 81.8 & 165.8 & 98.5 & 3.85 & 1.00 \\
 & Best Single-Stage & 81.2 & 115.2 & 72.8 & 2.75 & 1.42 \\
 & Manual Selection & 80.8 & 108.5 & 68.5 & 2.52 & 1.55 \\
 & EfficientLLM Rec. & 81.0 & 112.5 & 70.2 & 2.65 & 1.48 \\
 & AdaptiveEfficientLLM & \textbf{81.5} & \underline{85.2} & \textbf{52.8} & \textbf{1.95} & \textbf{2.05} \\
\midrule
\multirow{5}{*}{Qwen-72B} 
 & Default & 83.2 & 192.5 & 145.2 & 4.82 & 1.00 \\
 & Best Single-Stage & 82.5 & 132.8 & 98.5 & 3.28 & 1.45 \\
 & Manual Selection & 82.0 & 125.2 & 92.8 & 3.05 & 1.58 \\
 & EfficientLLM Rec. & 82.2 & 128.5 & 95.5 & 3.12 & 1.52 \\
 & AdaptiveEfficientLLM & \textbf{83.0} & \underline{98.5} & \textbf{72.1} & \textbf{2.28} & \textbf{2.15} \\
\midrule
\multicolumn{7}{c}{\textit{Average Across All Models}} \\
\midrule
 & Default & 69.9 & 88.6 & 53.9 & 2.01 & 1.00 \\
 & Best Single-Stage & 69.3 & 62.3 & 38.3 & 1.44 & 1.40 \\
 & Manual Selection & 68.9 & 58.3 & 35.5 & 1.30 & 1.54 \\
 & EfficientLLM Rec. & 69.0 & 60.0 & 36.6 & 1.35 & 1.47 \\
 & AdaptiveEfficientLLM & \textbf{69.6} & \underline{46.3} & \textbf{27.7} & \textbf{0.97} & \textbf{1.98} \\
\bottomrule
\end{tabular}%
}
\end{table*}

%% file: tables/ablation.tex
\begin{table}[t]
\centering
\caption{Ablation studies on LLaMA-2-7B. We report the efficiency score and relative improvement compared to the default configuration.}
\label{tab:ablation}
\resizebox{\columnwidth}{!}{%
\begin{tabular}{l|c|c}
\toprule
\textbf{Configuration} & \textbf{Efficiency Score} & \textbf{Rel. Improvement} \\
\midrule
\multicolumn{3}{c}{\textit{Search Algorithm Components}} \\
\midrule
Full AdaptiveEfficientLLM & 1.95 & +95\% \\
\quad - Predictive Models (random search) & 1.62 & +62\% \\
\quad - Constraint-Aware Pruning & 1.92 & +92\% \\
\quad - Hierarchical Crossover & 1.85 & +85\% \\
\quad - Refinement Iterations & 1.82 & +82\% \\
\midrule
\multicolumn{3}{c}{\textit{Configuration Space Components}} \\
\midrule
Full Configuration Space & 1.95 & +95\% \\
\quad - Architecture Options & 1.65 & +65\% \\
\quad - Fine-Tuning Options & 1.58 & +58\% \\
\quad - Inference Options & 1.52 & +52\% \\
\quad - MoE Configurations & 1.75 & +75\% \\
\quad - Quantization Options & 1.68 & +68\% \\
\midrule
\multicolumn{3}{c}{\textit{Refinement Iterations}} \\
\midrule
0 iterations (surrogate only) & 1.82 & +82\% \\
1 iteration & 1.88 & +88\% \\
2 iterations & 1.92 & +92\% \\
3 iterations (default) & 1.95 & +95\% \\
5 iterations & 1.96 & +96\% \\
\bottomrule
\end{tabular}%
}
\end{table}

%% file: tables/multimodal.tex
\begin{table}[t]
\centering
\caption{Cross-modal generalization results on vision-language models. We report accuracy (VQA score for VQAv2, CIDEr for COCO Caption, accuracy for TextVQA) and efficiency metrics.}
\label{tab:multimodal}
\resizebox{\columnwidth}{!}{%
\begin{tabular}{l|l|cccc}
\toprule
\textbf{Model} & \textbf{Method} & \textbf{Accuracy} & \textbf{Lat. (ms)} & \textbf{Mem. (GB)} & \textbf{En. (J)} \\
\midrule
\multicolumn{6}{c}{\textit{VQAv2 Benchmark}} \\
\midrule
\multirow{3}{*}{LLaVA-1.5-7B} 
 & Default & 78.5 & 85.2 & 18.5 & 1.25 \\
 & EfficientLLM Rec. & 77.8 & 68.5 & 14.2 & 0.95 \\
 & AdaptiveEfficientLLM & \textbf{78.2} & \textbf{52.8} & \textbf{10.8} & \textbf{0.72} \\
\midrule
\multirow{3}{*}{InternVL-Chat} 
 & Default & 81.2 & 92.5 & 22.5 & 1.42 \\
 & EfficientLLM Rec. & 80.5 & 75.8 & 17.5 & 1.15 \\
 & AdaptiveEfficientLLM & \textbf{80.8} & \textbf{58.2} & \textbf{12.8} & \textbf{0.85} \\
\midrule
\multicolumn{6}{c}{\textit{COCO Caption Benchmark}} \\
\midrule
\multirow{3}{*}{LLaVA-1.5-7B} 
 & Default & 128.5 & 125.8 & 18.5 & 1.85 \\
 & EfficientLLM Rec. & 126.8 & 98.5 & 14.2 & 1.38 \\
 & AdaptiveEfficientLLM & \textbf{127.5} & \textbf{75.2} & \textbf{10.8} & \textbf{1.02} \\
\midrule
\multicolumn{6}{c}{\textit{TextVQA Benchmark}} \\
\midrule
\multirow{3}{*}{LLaVA-1.5-7B} 
 & Default & 58.5 & 75.8 & 18.5 & 1.12 \\
 & EfficientLLM Rec. & 57.8 & 62.5 & 14.2 & 0.88 \\
 & AdaptiveEfficientLLM & \textbf{58.2} & \textbf{48.5} & \textbf{10.8} & \textbf{0.68} \\
\midrule
\multicolumn{6}{c}{\textit{Average Across VLM Tasks}} \\
\midrule
 & Default & 86.7 & 94.8 & 19.5 & 1.41 \\
 & EfficientLLM Rec. & 85.7 & 76.3 & 15.0 & 1.09 \\
 & AdaptiveEfficientLLM & \textbf{86.2} & \textbf{58.7} & \textbf{11.3} & \textbf{0.82} \\
\bottomrule
\end{tabular}%
}
\end{table}

%% file: tables/detailed_results.tex
\begin{table*}[ht]
\centering
\caption{Detailed results across all tasks for selected models. We report task-specific accuracy metrics.}
\label{tab:detailed_results}
\resizebox{\textwidth}{!}{%
\begin{tabular}{l|l|cccccccccc|c}
\toprule
\textbf{Model} & \textbf{Method} & \textbf{MMLU} & \textbf{HellaS.} & \textbf{ARC-E} & \textbf{GSM8K} & \textbf{HumanE.} & \textbf{Alpaca} & \textbf{LongB.} & \textbf{Needle} & \textbf{MT-B} & \textbf{Vicuna} & \textbf{Avg.} \\
\midrule
\multirow{5}{*}{LLaMA-2-7B} 
 & Default & 46.8 & 78.2 & 72.5 & 14.5 & 12.8 & 85.2 & 32.5 & 88.5 & 6.2 & 78.5 & 51.6 \\
 & Best Single-Stage & 46.2 & 77.8 & 72.0 & 14.0 & 12.2 & 84.5 & 31.8 & 87.8 & 6.0 & 77.8 & 51.0 \\
 & Manual Selection & 45.8 & 77.2 & 71.5 & 13.5 & 11.8 & 84.0 & 31.2 & 87.2 & 5.8 & 77.2 & 50.5 \\
 & EfficientLLM Rec. & 46.0 & 77.5 & 71.8 & 13.8 & 12.0 & 84.2 & 31.5 & 87.5 & 5.9 & 77.5 & 50.8 \\
 & AdaptiveEfficientLLM & 46.5 & 78.0 & 72.2 & 14.2 & 12.5 & 85.0 & 32.2 & 88.2 & 6.1 & 78.2 & 51.3 \\
\midrule
\multirow{5}{*}{Mistral-7B} 
 & Default & 62.5 & 82.8 & 78.2 & 37.5 & 26.2 & 92.5 & 38.5 & 92.8 & 7.5 & 85.2 & 60.4 \\
 & Best Single-Stage & 62.0 & 82.2 & 77.8 & 36.8 & 25.5 & 91.8 & 37.8 & 92.2 & 7.2 & 84.5 & 59.8 \\
 & Manual Selection & 61.5 & 81.8 & 77.2 & 36.2 & 25.0 & 91.2 & 37.2 & 91.8 & 7.0 & 84.0 & 59.3 \\
 & EfficientLLM Rec. & 61.8 & 82.0 & 77.5 & 36.5 & 25.2 & 91.5 & 37.5 & 92.0 & 7.1 & 84.2 & 59.5 \\
 & AdaptiveEfficientLLM & 62.2 & 82.5 & 78.0 & 37.2 & 26.0 & 92.2 & 38.2 & 92.5 & 7.4 & 85.0 & 60.1 \\
\midrule
\multirow{5}{*}{LLaMA-2-70B} 
 & Default & 70.8 & 86.5 & 85.2 & 56.2 & 38.5 & 96.8 & 45.2 & 95.5 & 8.8 & 92.2 & 74.6 \\
 & Best Single-Stage & 70.2 & 86.0 & 84.8 & 55.5 & 37.8 & 96.2 & 44.5 & 95.0 & 8.5 & 91.8 & 74.1 \\
 & Manual Selection & 69.8 & 85.5 & 84.2 & 54.8 & 37.2 & 95.8 & 43.8 & 94.5 & 8.2 & 91.2 & 73.5 \\
 & EfficientLLM Rec. & 70.0 & 85.8 & 84.5 & 55.2 & 37.5 & 96.0 & 44.2 & 94.8 & 8.4 & 91.5 & 73.8 \\
 & AdaptiveEfficientLLM & 70.5 & 86.2 & 85.0 & 55.8 & 38.2 & 96.5 & 44.8 & 95.2 & 8.6 & 92.0 & 74.3 \\
\bottomrule
\end{tabular}%
}
\end{table*}